\documentclass[letterpaper, 10 pt, journal, twoside]{IEEEtran}
\pdfoutput=1
\IEEEoverridecommandlockouts               
\usepackage{multirow}
\usepackage{color}
\usepackage{xcolor}
\usepackage{url}
\usepackage{cite}
\usepackage{amsmath,amssymb,amsfonts}
\usepackage{algorithmic}
\usepackage{graphicx}
\usepackage{caption}
\usepackage{float}
\usepackage{textcomp}
\usepackage{tensor}
\usepackage{diagbox}
\usepackage{booktabs}
\usepackage{wrapfig}
\usepackage{subfigure}
\usepackage{multirow}
\usepackage{verbatim}
\usepackage{diagbox}

\begin{document}

 \title{Design and Control of a Highly Redundant Rigid–Flexible Coupling Robot to Assist the COVID-19 Oropharyngeal-Swab Sampling}

\author{ Yingbai Hu$^{3,2 \ddagger}$, Jian Li$^{1,2 \ddagger}$, Yongquan Chen$^{1,2*}$, Qiwen Wang$^{2,1}$, Chuliang Chi$^{2,1}$, Heng Zhang$^{2,1}$, Qing Gao$^{2,1}$, \\Yuanmin Lan$^{6,2}$, Zheng Li$^{4,2}$, Zonggao Mu$^{5,2}$, Zhenglong Sun$^{1,2}$, Alois Knoll$^3$

\thanks{* This work was supported by The Chinese University of Hong Kong, Shenzhen, and the Shenzhen Institute of Artificial Intelligence and Robotics for Society. 
\begin{itshape}(Corresponding author: Yongquan Chen, Email: yqchen@cuhk.edu.cn)\end{itshape}
}

\thanks{$^{\ddagger}$ These authors contributed equally to this work.}
\thanks{$^{1}$ Robotics and Intelligent Manufacturing \& School of Science and Engineering, The Chinese University of Hong Kong, Shenzhen, 518172, China.}
\thanks{$^{2}$ Shenzhen Institute of Artificial Intelligence and Robotics for Society, 518129, China.}
\thanks{$^{3}$ Chair of Robotics, Artificial Intelligence and Real-time Systems, Technische Universit\"at M\"unchen, M\"unchen, 85748, Germany.}%
\thanks{$^{4}$ Department of surgery, and Chow Yuk Ho Technology Centre for Innovative Medicine, The Chinese University of Hong Kong, Hong Kong.}%
\thanks{$^{5}$ School of Mechanical Engineering, Shandong University of Technology, Zibo, 255000, China.}%
\thanks{$^{6}$ Longgang District People's Hospital of Shenzhen, 518172, China.}%
}

\markboth{}
{ \MakeLowercase{\textit{}} } 

\maketitle


\begin{abstract}
The outbreak of novel coronavirus pneumonia (COVID-19) has caused mortality and morbidity worldwide.
Oropharyngeal-swab (OP-swab) sampling is widely used  for the diagnosis of COVID-19 in the world.
To avoid the clinical staff from being affected by the virus, we developed a 9-degree-of-freedom (DOF) rigid–flexible coupling (RFC) robot to assist the COVID-19 OP-swab sampling.
This robot is composed of a visual system, UR5 robot arm, micro-pneumatic actuator and force-sensing system. The robot is expected to reduce risk and free up the clinical staff from the long-term repetitive sampling work.
Compared with a rigid sampling robot, the developed force-sensing RFC robot can facilitate OP-swab sampling procedures in a safer and softer way.
In addition, a  varying-parameter zeroing neural network-based optimization method is also proposed for motion planning of the 9-DOF redundant manipulator.
The developed robot system is validated by OP-swab sampling on both oral cavity phantoms and volunteers.
\end{abstract}

\IEEEpeerreviewmaketitle	

\section{Introduction}
\IEEEPARstart{T}{he} outbreak of novel coronavirus pneumonia (COVID-19) is affecting the entire world, which has caused a large number of deaths with an increase in the spread of COVID-19.
To control the spread of COVID-19 at the early stage, oropharyngeal-swab (OP-swab) sampling is commonly adopted with respect to sample collection and specimen sources for diagnosis \cite{wang2020comparison}.
 \begin{figure}[htp!]
	\centering{		\includegraphics[width=1.00\linewidth, height=0.84 \linewidth]{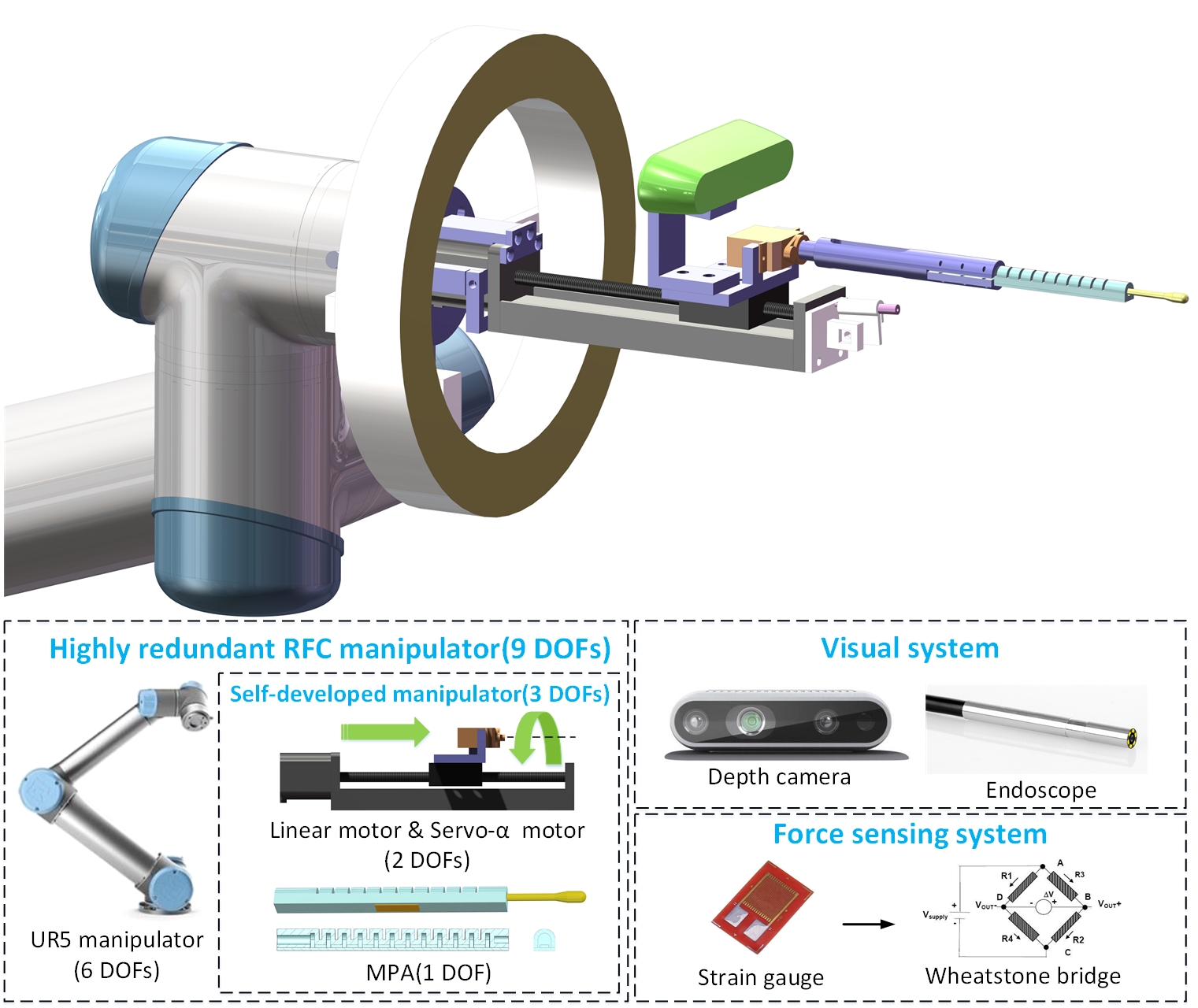}}
	\caption{ 9-degree-of-freedom (DOF)  redundant rigid–flexible coupling robot }
	\label{fig.9DOFs_RFC_Robot}
\end{figure}
However, protecting the safety of medical staff during the sampling process raises a new challenge because of susceptibility to infection from person to person through respiratory droplets and contact transmission in an unprotected environment \cite{xu2020open}. 
A variety of related studies have reported that respiratory droplets, feces and urine are the routes of transmission \cite{hindson2020covid}.
To address these issues in OP-swab sampling,  robotics could play a key role in disease prevention.

\begin{figure*}[htp!]
	\centering{
	\includegraphics[width=0.99\linewidth]{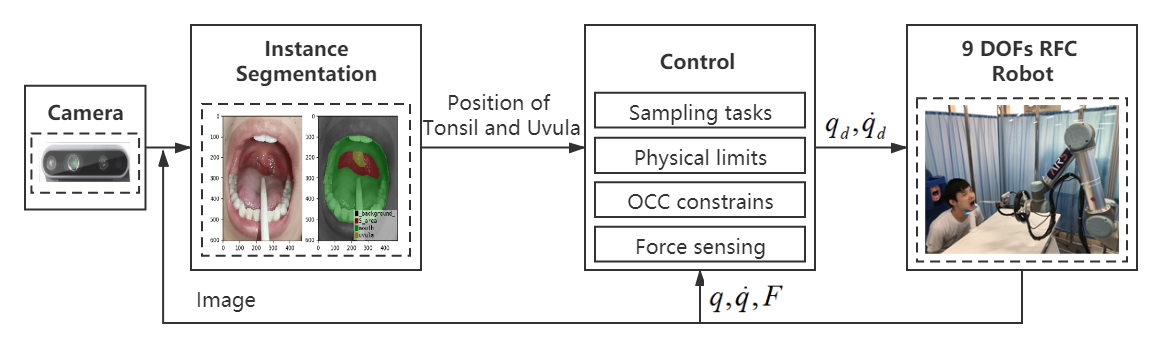}}
	\caption{Control framework of COVID-19 OP-swab sampling with a redundant rigid–flexible coupling robot}
	\label{fig.block diagram}
\end{figure*}

Considering the high risk of infection of COVID-19 and negative tests of nasopharyngeal swabs (NP-swab) caused by irregular sampling,  it is necessary to design an OP-swab sampling robot to assist healthcare staff through remote access.
Robot-assisted OP-swab sampling is a promising technique because it relieves the burden from medical staff, is convenient for large-scale deployment, is cost-effective, and offers sampling standardization.
As reported in \cite{yang2020combating}, an OP-swab robot could speed up the sampling process especially because there is a lack of qualified healthcare workers. 
In \cite{li2020clinical} and \cite{liu2020},  
a semi-automatic OP-swab robot (compared to the traditional human method) was developed with a teleoperation scheme that achieved good performance in clinical practice with a 95\% success rate.
The OP-swab robot system mainly includes a snake-like manipulator, two 
haptic devices(Omega.3 and Omega.6) for teleoperation, an endoscope for visualization (assistance operating in the oral cavity) and a force-sensing system for safety protection.
All OP-swab sampling processes were controlled by experts' teleoperation, where one haptic device (Omega.3) control the tongue depressor and another haptic device (Omega.6) operate the manipulator.
Wang {\it{et al.}}~ \cite{wang2020design} have designed a low-cost robot for assistance in sampling of NP-swab; a swab gripper is attached to the rotation link with its extruded active 2-degree-of-freedom (DOF) end-effector for actuating the swab and a generic 6-DOF passive arm for global positioning.
In \cite{Danmark2020}, the team from the University of Southern Denmark and the Lifeline Robotics company developed the first automatic swab robot for COVID-19 sampling.
In \cite{Danmark2020}, the robot is integrated with a UR5 manipulator, a visual system and dexterous rotatable rigid connectors assembled with  OP-swab which could perform the sampling task quickly using an automated scheme.

Of note, medical security remains vital to OP-swab sampling because human throat is fragile and easily injured.
Nonetheless, although many desirable results were obtained, the abovementioned end-effectors were designed on the basis of a rigid body structure, which may cause physical injuries of the oral cavity during improper operation or other medical malpractices. 
On the basis of our previous work \cite{li2015novel}, a novel micro-pneumatic  actuator (MPA) for throat swab sampling is developed to achieve flexible collection  and integrate a force sensor, which provides safe, stable, and reliable sampling experience, as shown in Fig. \ref{fig.9DOFs_RFC_Robot}. 
Furthermore, force-sensing actuation offers compliance, which is helpful
against shocks, particularly during interaction with oral cavity.
Compared with the rigid body-based sampling robots \cite{li2020clinical, wang2020design} and \cite{Danmark2020}, 
the developed MPA is made of soft material, which is safer and lighter.
Moreover, MPA is smaller and has a 7.5 mm cross-sectional diameter, which is convenient for working in human oral cavity.

To avoid contact with the OP-swab specimen and collision with the manipulator, it is essential to consider the constraints of the oral cavity space during motion planning.
Inspired by our previous work \cite{su2020improved} on remote center of motion (RCM) techniques in minimally invasive surgical robots,  the second to the last link of the OP-swab sampling robot is constrained with the oral cavity center (OCC) constraint, which guarantees the absence of collision between the inserted end-effector with the oral cavity.
Although mechanical implementation is usually safer for an open  minimally invasive surgery, programmable RCM is cost effective and convenient to be implemented 
 \cite{su2019improved}.
The traditional Cartesian adaptive control in \cite{su2019improved} was applied for surgical control with the RCM constraint but without considering the physical limits of the manipulator, such as the joint angle and velocity.
For practical applications, we propose an optimization strategy for the multi-constraint problem, where the joint angle and velocity, and the OCC constraint-based kinematic model are derived as a quadratic programming problem. 
In this study, a comparison between the varying-parameter zeroing neural network (VP-ZNN) and the traditional gradient descent method \cite{mirrazavi2018unified} is proposed for the optimization problem that has achieved superior performance, which can converge at the finite time owing to the optimal solution changing over time.

To test the developed OP-swab sampling robot, which is shown in Fig.~\ref{fig.9DOFs_RFC_Robot}, the experiments are simultaneously conducted in both the human  oral  cavity  phantom  and  volunteers, regarding the sampling time, operation complexity for medical staff, safety and effectiveness. The framework of COVID-19 OP-swab sampling using the RFC robot is shown in Fig.~\ref{fig.block diagram}.
The contributions of this study can be summarized as follows.
\begin{enumerate}
\item A rigid–flexible coupling robot is designed to assist COVID-19 OP-swab sampling, where the MPA for OP-swab sampling is developed to achieve flexible sampling and thereby provides compliant, safe, stable, and reliable sampling.
    
\item An improved motion planning method is  proposed for the 9-DOF highly redundant robot-assisted OP-swab sampling system, which guarantees efficiency and accuracy,  where the multi-constraint kinematics model is derived as an optimization problem; then, the VP-ZNN is applied to solve the optimization problem.
    
\item  Substantive  practical experiment testing of the developed OP-swab sampling robot system is demonstrated in the human oral cavity phantom and volunteers.
\end{enumerate}


\section{Design of the  Rigid–Flexible Coupling Robot}
\label{sec.design}

\subsection{ Redundant Rigid–Flexible Coupling Robot}

The rigid–flexible coupling robot for OP-swab sampling consists of a UR5 manipulator and a self-developed rigid–flexible coupling manipulator. The overall structure of the 9-DOF redundant rigid–flexible coupling robot is shown in  Fig.~\ref{fig.9DOFs_RFC_Robot}.  The self-developed manipulator has 3 DOFs, including a linear motor (prismatic joint), a servo–$\alpha$ motor (revolute joint), and an MPA that can change the offset distance of throat swab from the center.

 \begin{figure}[htp!]
	\centering{
		\includegraphics[width=1.0\linewidth, height=0.99 \linewidth]{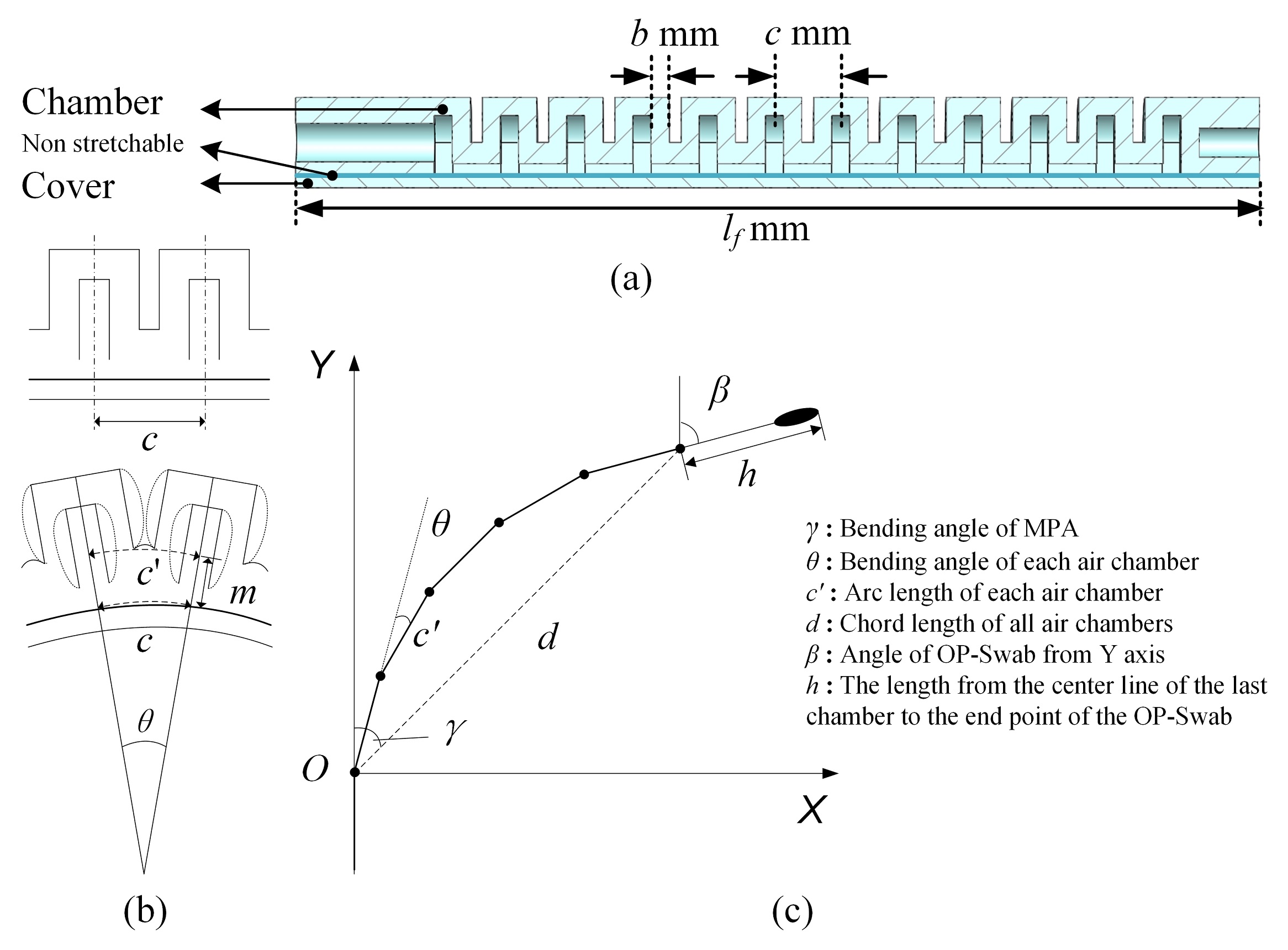}}
	\caption{(a) Cross section of the micro-pneumatic actuator(MPA). (b) bending process of MPA. and (c) 2D modeling of MPA.}
	\label{fig.Bending_of_fingers}
\end{figure}

  \begin{figure}
\flushleft
\subfigure[]{
\includegraphics[width=2.8cm,height=4cm]{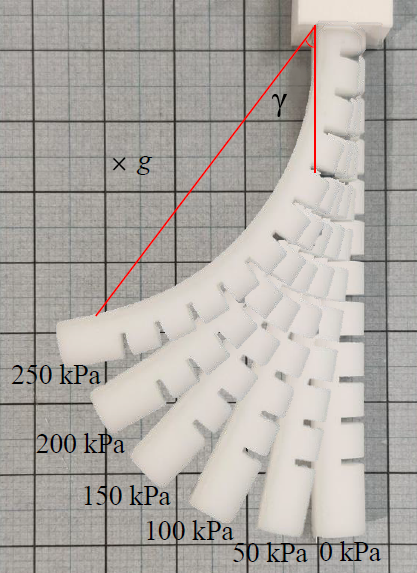}
}
\quad
\hspace{-8mm}
\subfigure[]{
\includegraphics[width=5.4cm,height=4cm]{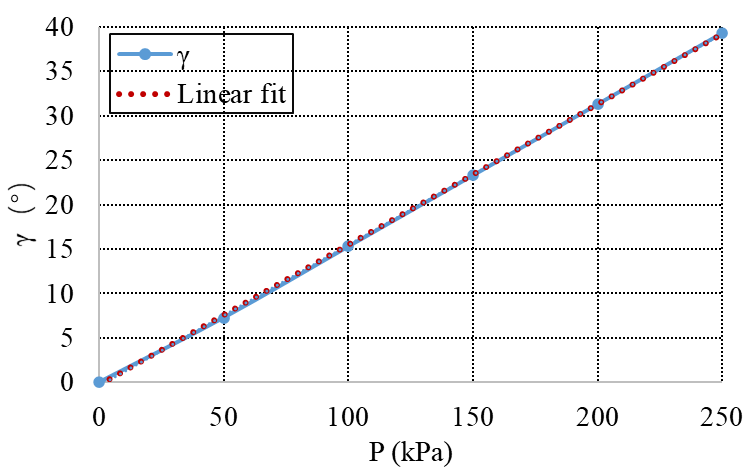}
}
\caption{
Experimental results of  MPA. (a) Deformation of MPA with a 50-shore hardness under different air pressures. (b) Relationship between bending angle $\gamma$ and air pressure}
\label{fig.cali}
\end{figure}


\subsection{Design, Modeling, and Fabrication of MPA}
The design of MPA refers to the wrinkle shape of the elastomer robot \cite{wang2017prestressed}. Figure \ref{fig.Bending_of_fingers} shows that MPA consists of $ n $ effective air chambers.  The sectional view of MPA is shown in Fig.~\ref{fig.Bending_of_fingers}(a), which contains a chamber part and a cover part. These two parts are bonded together to form a single-DOF MPA. $ b $ is the wall thickness of the air chamber; $ c $  is the distance between the air chambers, and  $ l_f $ is the total length of  MPA.  When MPA is inflated, the air chambers will expand and repel each other, which causes MPA to bend. The bending process of MPA is shown in Fig.~\ref{fig.Bending_of_fingers}(b). 

Assuming that all air chambers have the same bending under the same pressure, it is easy to obtain the geometric relation:
\begin{equation}
 	\label{equ-gama}	
\theta {\rm{ = }}\beta /n {\rm{ = 2}}\gamma /n
\end{equation}

Owing to the small value of $\theta$ after bending, the chord length is approximately equal to the arc length at the non-tensile layer. The distance between the center line and the non-tensile layer is $ m $, and the arc length $ c' $ is also approximately equal to the chord length; thus, $ c' $ is 
\begin{equation}
 	\label{equ1}
c' = c + 2m\sin (\theta /2)
\end{equation}
The overall motion model is shown in Fig.~\ref{fig.Bending_of_fingers} (c), where the chord length $ d $ is
\begin{equation}
 	\label{equ2}	
d = \frac{{c'\sin ({n}\theta /2)}}{{\sin (\theta /2)}}
\end{equation}
The forward kinematics of MPA is modeled as
\begin{equation}
 \left\{ {\begin{array}{*{20}{c}}
{{x} = d\sin \left( {n\frac{\theta }{2}} \right){\rm{ + }}h\sin \left( \beta  \right)}\\
~\\{{y} = d\cos \left( {n\frac{\theta }{2}} \right){\rm{ + }}h\cos \left( \beta  \right)}
\end{array}} \right. 
\label{equ3}
\end{equation}

 According to the abovementioned information, we design MPA with 12 air chambers.
 However, the number of effective air chambers, i.e., the number of effective joints is $ n=11 $ because only half of the head chamber and tail chamber will bend.  The wall thickness  $ b $ of each chamber is 1.5 mm , the width of each air chamber $ c $  is 4.5 mm and the total length $ l_f $ is 80 mm.

The chamber part and the cover part of MPA are made by 3D printing with a 50-shore hardness.
The air pressure is controlled by the proportional valve ITV1050-312L. Then, a deformation experiment of MPA was conducted, and the experimental results are shown in Fig.~\ref{fig.cali}(a), where the gravity is directly perpendicularly to the plane of the paper.
The relationship between the bending angle $\gamma$ and air pressure is expressed in Fig.~\ref{fig.cali}(b).
It is observed that the bending angle $\gamma$ is approximately proportional to the air pressure. 
After data fitting, the linear equation for the angle $\gamma$ ($^{\circ}$) and pressure $P$ (kPa) is written as
\begin{equation}
	\label{equ5}	
 \gamma {\rm{ = 0}}.160325P{\rm{ - 0}}.731885 
\end{equation}
The correlation coefficient $ R $ is 0.99990 and the standard deviation $ S $ is 0.6464, which confirms that $\gamma$ and $P$ are linearly dependent. By combining equations (\ref{equ-gama}) and (\ref{equ5}), we obtain
\begin{equation}
	\label{equ6}	
 \theta {\rm{ = 0}}.02915P{\rm{ - 0}}.13307 
\end{equation}
Finally, the end-point position of MPA can be calculated by combining equations (\ref{equ1}), (\ref{equ2}), (\ref{equ3}), and (\ref{equ6}).

\begin{figure}[htp!]
	\centering{
		\includegraphics[width=0.99\linewidth]{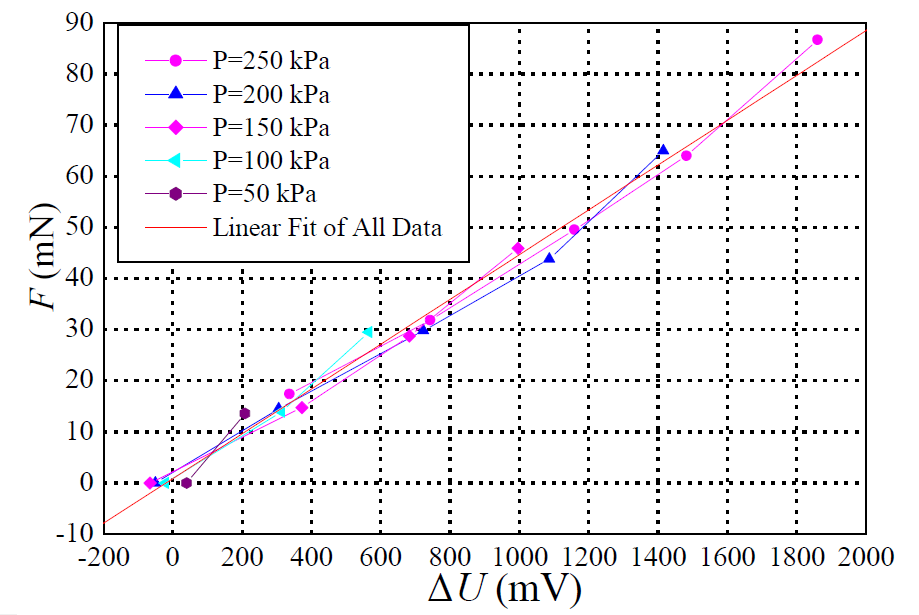}}
    	\caption{Relationship  between $ F $ and $ \Delta U $ }
	\label{fig.relationshipUF}
\end{figure}
\subsection{Force Sensing}
To detect the contact force $ F $ during OP-swab sampling, a strain gauge sensor (BF350-3AA23T0) attached to MPA is adopted. It is necessary to explore the mapping relationship between the force and voltage for force measurement calibration.
Actually, the current output voltage $ U_{now} $ is related to two factors, i.e., voltage $ \Delta U $ generated by the force, and voltage $ {U_{F{\rm{ = 0}}}} $ caused by the air pressure without a load. 
Therefore, we have $\Delta U{\rm{ = }}{U_{{\rm{now}}}} - {U_{F{\rm{ = 0}}}}$. After 20 group experiments, the relationship between no-load voltage and air pressure is obtained by linear fitting:  $ {U_{F = 0}}{\rm{ = }} - 10.693P + 2770.571 $  (mV). Force calibration was performed using the electronic balance (CX-668). The relationship between $ \Delta U $ and $ F $ is approximately linear; thus, we know that \begin{equation}
 F = {\rm{0}}{\rm{.04378}}\Delta U + {\rm{0}}{\rm{.88828}} 
\end{equation}
The details are shown in Fig.~\ref{fig.relationshipUF}.
The correlation coefficient $ R $ is 0.99461 and the standard deviation $ S $ is 2.65733. 


\section{Optimization Control for the 9-DOF Redundant Manipulator}
\label{sec.control}
\subsection{Manipulator Kinematics Model with OCC Constraints}
\begin{figure}[htp!]
	\centering{
		\includegraphics[width=0.99\linewidth]{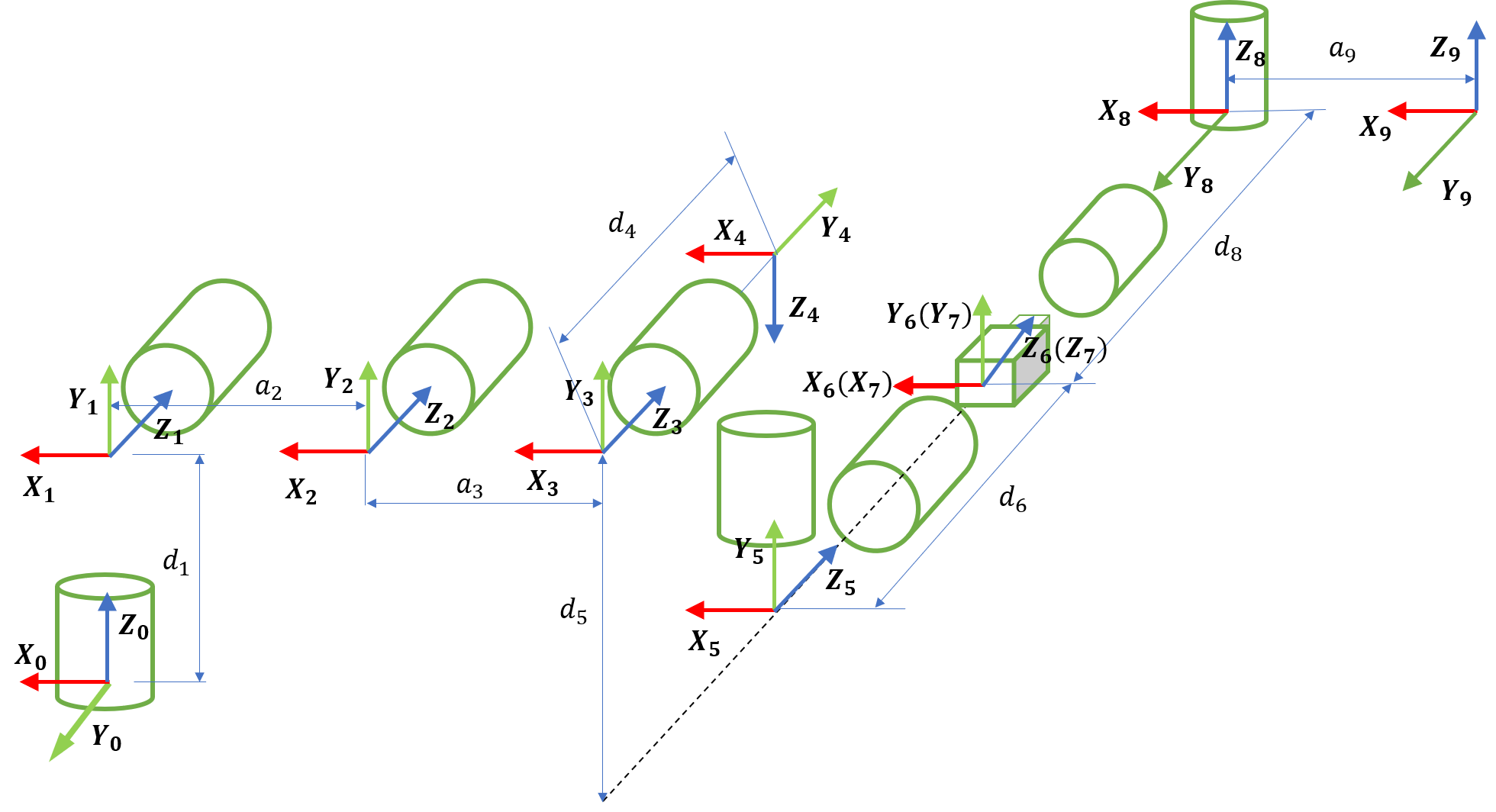}}
	\caption{Coordinate system of the 9-DOF redundant RFC manipulator.}
	\label{fig.coordinatesystem}
\end{figure}

The coordinate system of the 9-DOF redundant RFC manipulator is shown in Fig.~\ref{fig.coordinatesystem}. The forward kinematics model \cite{li2017asymmetric, li2018adaptive} of the 9-DOF RFC robot can be defined as follows:
\begin{equation}
{J}\dot q = {\dot l_{d}}
\label{eq.Jacobian}
\end{equation}
where $J \in \mathbb{R}^{m \times n}$ denotes the Jacobian matrix.
The joint angle and velocity constraints are defined as follows:
\begin{alignat}{2}
{q_i ^ - } \le q_i  \le {q_i ^ + } \label{eq.limit0}\\
{{\dot q_i }^ - } \le \dot q_i  \le {{\dot q_i }^ + }
\label{eq.limit1}
\end{alignat}
where $q_i^-$ and $q_i^ +$ present the lower and upper bounds of $q_i$, respectively; and $\dot q_i ^ -$ and $\dot q_i ^ +$ denote the lower and upper bounds of the joint velocity $\dot q$, respectively.

Actually, the constraints in (\ref{eq.limit0}) can be converted as
\begin{equation}\label{eq.limit2}
\sigma \left( {{q_i ^ - } - q_i } \right) \le \dot q_i  \le \sigma \left( {{q_i ^ + } - q_i } \right)
\end{equation}
where $\sigma$ is the positive constant.
Therefore,  according to  (\ref{eq.limit1})  and (\ref{eq.limit2}), the joint angle and velocity constraints can be rewritten as a new constraint in the velocity level:
\begin{alignat}{2}
&{\rho_i ^ - } \le \dot q_i  \le {\rho_i ^ + } \label{eq.limit3}\\
&\rho _i^ -  = \max \left\{ {\dot q _i^ - ,\sigma \left( {q _i^ -  - {q _i}} \right)} \right\} \nonumber \\
&{\rho_i ^ + } = \min \left\{ {\dot q _i^ + ,\sigma \left( {q _i^ +  - {q _i}} \right)} \right\} \nonumber
\end{alignat}

Because the RFC robot is a 9-DOF highly redundant manipulator, there are infinite solutions of $\dot q$ in (\ref{eq.Jacobian}) by the inverse kinematics.
However, the convergence rate, accuracy, and computational complexity of the pseudoinverse-type solution in inverse kinematics cannot satisfy the requirements.
Consequently, we need to identify an optimization solution under multiple constraints; thus, the inverse kinematics problem can be expressed as a new optimization problem.
The first-priority optimization problem associated with OP-swab sampling is expressed as follows:
\begin{alignat}{2}
\text{min}~~~&\frac{1}{2}{\dot{q} ^T}W\dot{q} \label{eq.qpmin1} \\
\text{s.t.}~~&{J\left(q\right)}\dot{q} = {\dot{l}_{d}}\label{eq.qpst1} \\
&{\rho ^ - }\le {\dot{q }} \le &{\rho ^ + }\label{eq.qpstq1}
\end{alignat}
where $\dot{l}_{d}$ represents the reference velocity associated with  OP-swab  sampling tasks. The weight matrix $M$ is set as the identity matrix.

\begin{figure}[htp!]
	\centering{
		\includegraphics[width=0.9\linewidth]{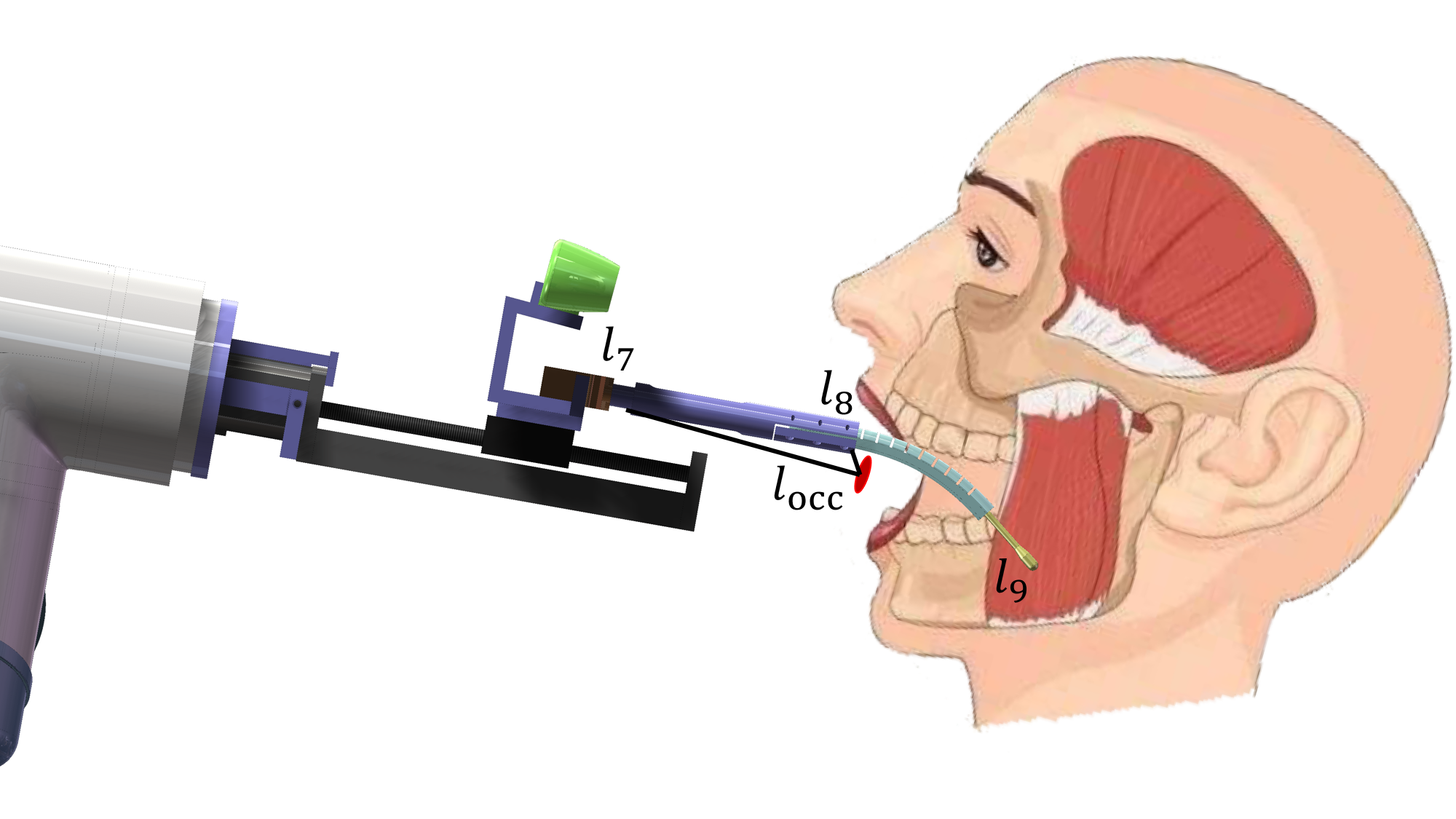}}
	\caption{Constraint with the oral cavity center.}
	\label{fig.constraintRCM}
\end{figure}

Considering the OCC constraint, the link-$L_{n-1}$ of the RFC manipulator needs to pass through the OCC, and $L_n$ performs the sampling tasks, which is different from the traditional RCM constraint in the last link. The geometric relationship is shown in Fig.~\ref{fig.constraintRCM}.
For the $n$-DOF OP-swab sampling robot, the forward kinematics mapping function of Cartesian position $ l_{n-2} \in \mathbb{R}^m$ and $l_{n-1} \in \mathbb{R}^m$ can be defined as follows:
\begin{equation}
\begin{aligned}
&{{l}_{n-2}=f_{n-2}( q)} \\ 
&{{l}_{n-1}=f_{n-1}( q)}
\end{aligned}
\label {eq2}
\end{equation}

Unlike common RCM constraints, 
 $l_{occ}$ should be always on the straight line straight line between $l_{n-1}$ and $l_{n-2}$ (second to the last link), where  $l_{n-1}$ is the end position of the link $L_{n-1}$ and $l_{n-2}$ is the  end position of the link $L_{n-2}$. During the  actual OP-swab sampling, we want to keep the error of the OCC constraint $E_{occ}$ as small as possible. Lines 1 and 2 are constructed as follows: ${\overrightarrow{l_{n-2}r_{n-1}}}=l_{n-1}-l_{n-2}$, ${\overrightarrow{l_{n-2} l_{occ}}}=l_{occ}-l_{n-2}$, respectively.
Utilizing the relationship between $E_{occ}$ and the vector projection, $E_{occ}$ can be further written as follows:
\begin{alignat}{2}
{E_{occ}} &= \frac{{\overrightarrow {{l_{n-2}}{l_{occ}}}  \times \overrightarrow {{l_{n-2}}{l_{n-1}}} }}{L}
\label{eq.OCCerror}
\end{alignat}
where $L=\left\| {{l_{n-2}} - {l_{n-1}}} \right\|$ is the length of the second to the last link.
The derivative of $E_{occ}$ in (\ref{eq.OCCerror}) with respect to time is reformulated as follows:
\begin{equation}\label{eq.OCC_constrant}
J_{occ}\dot q= \dot{E}_{occ}
\end{equation}
where $J_{occ} \in \mathbb{R}^{m \times n}$ denotes the Jacobian matrix of the OCC constraint.
Regarding the OCC constraint task, we want to keep the OCC error $E_{occ}$ at the minimum value.

In this section, it is necessary to find a feasible solution for multiple constraints and guarantee that the tracking error and OCC error always remain in a small range. For COVID-19 sampling tasks, we aim to reformulate OCC (\ref{eq.OCC_constrant}) and joint physical limits (\ref{eq.qpstq1}) in an optimization scheme and design a method to solve the optimization problem. Consequently, by simultaneously taking OCC, end-effector task, and joint physical constraints into account, the new multi-task optimization problem can be formulated as

\begin{alignat}{2}
\text{min} ~~~&\frac{1}{2}{{\dot q }^T}W\dot q \label{eq.multi-opt1}\\
\text{s.t.} ~~~&J\dot q  = v_d\nonumber \\
&{J_{occ}}\dot q  = v_{occ} \nonumber \\
&{\rho ^ - } \le \dot q  \le {\rho ^ + } \nonumber
\end{alignat}
where $v_d={{\dot l}_{nd}}$ and $v_{occ}=0$.
Without the explicit expression of $l_n$ and $e_{occ}$, the actual trajectory will drift and the position error cannot converge to zero from a random initial position. 
To overcome this issue in (\ref{eq.multi-opt1}), a feedback item associated with position signals is integrated into $v_d$ and $v_{occ}$:
\begin{alignat}{2}
&v_d  =  - {k_1}\left( {{f_n}\left( q  \right) - {l_{nd}}} \right) + {{\dot l}_{nd}}\\
&v_{occ}  =  - {k_2}\left( {{l_{occ}}} \right)
\end{alignat}
Of note, we should manage the priority strategy for the multiple-task optimization problem by different weights. Therefore, the objective function is defined as
\begin{equation}
   F(\dot q)= \frac{{{\varepsilon_0}}}{2}{{\dot q }^T}\dot q  + \frac{{{\varepsilon_1}}}{2}{\left\| {J\dot q  - v_d} \right\|^2} + \frac{{{\varepsilon_2}}}{2}{\left\| {{J_{occ}}\dot q-v_{occ} } \right\|^2} \label{eq.multi-opt2}
\end{equation}

The optimization problem in (\ref{eq.multi-opt2}) can be rewritten as follows:
\begin{alignat}{2}
\text{min} ~~~&F(\dot q) \\
\text{s.t.} ~~~&J\dot q  = v_d \nonumber \\
&{J_{occ}}\dot q  = v_{occ} \nonumber\\
&{\rho ^ - } \le \dot q  \le {\rho ^ + }\nonumber
\end{alignat}
where $\varepsilon_0>0$, $\varepsilon_1>0$ and $\varepsilon_2>0$ are the constants employed to prioritize different tasks.

\subsection{Neural Network Design}\label{sec.VP-ZNN}


In this section, we describe the design of the VP-ZNN, which is used to solve the optimization problem in (\ref{eq.multi-opt2}). First, the multiple-task optimization problem in (\ref{eq.multi-opt2}) is converted into an equivalent problem; thus, the VP-ZNN is employed to solve it.


The Lagrange function of (\ref{eq.multi-opt2}) constraints is formulated as follows:
\begin{alignat}{2}
&\mathcal{L}\left( {\dot q ,{\xi _1},{\xi _2}} \right) = \frac{{{\varepsilon_1}}}{2}{\left\| {J\dot q  - v_d} \right\|^2} + \frac{{{\varepsilon_2}}}{2}{\left\| {{J_{occ}}\dot q-v_{occ} } \right\|^2} \nonumber \\
& ~~~~~\frac{{{\varepsilon_0}}}{2}{{\dot q }^T}\dot q+ \xi _1^T\left( {{v_d} - J\dot q } \right) + \xi _2^T\left(v_{occ} { - {J_{occ}}\dot q } \right)
\label{eq.Lagrange}
\end{alignat}
where $\xi_1 \in \mathbb{R}^{m}$ and  $\xi_2 \in \mathbb{R}^{m}$;  $\nabla \mathcal{L} = {\left[ {\frac{{\partial \mathcal{L}}}{{\partial \dot q }},\frac{{\partial \mathcal{L}}}{{\partial {\xi _1}}},\frac{{\partial \mathcal{L}}}{{\partial {\xi_2}}}} \right]^T}$ indicates the gradient of (\ref{eq.Lagrange}).
As the KKT condition defined in \cite{ facchinei2014solving}, if $\nabla\mathcal{ L} $ is continuous, the optimization solution satisfies the following condition:
\begin{equation}\label{eq.Lagrangecondition}
\nabla \mathcal{L} = 0
\end{equation}
The state decision variable $l\left( t \right) = {\left[ {{{\dot q}},{{{\xi }}_1},{{{\xi }}_2}} \right]^T} \in R^{n+2m}$.
The problem in (\ref{eq.Lagrangecondition}) is equivalent to the following:
\begin{alignat}{2}
B(t)l(t)=P(t)
\end{alignat}
where $B\in \mathbb{R}^{(n+2m)\times (n+2m)}$ and  $P(t)\in \mathbb{R}^{(n+2m)}$.
The bounds of state variable $\dot q^b$  are expressed as 
\[\dot q^b= \left\{ \begin{array}{l}
{\rho ^ + },~~\dot q > {\rho ^ + }\\
\rho ,~~~{\rho ^ - } \le \dot q \le {\rho ^ + }\\
{\rho ^ - },~~\dot q < {\rho ^ - }
\end{array} \right.\]
Therefore, the bound of state variable $l(t)$ are defined as
\[{l^b}\left( t \right) = \left[ {\begin{array}{*{20}{c}}
{{{\dot q}^b}}&{\xi _1^b}&{\xi _2^b}
\end{array}} \right],~\xi _1^b,\xi _2^b \in \mathbb{R}\]
The error model of the novel VP-ZNN is expressed as
\begin{alignat}{2}
e\left( t \right) = B\left( t \right)l\left( t \right) - P\left( t \right)
\label{eq.novel VP-ZNNerror}
\end{alignat}
To ensure the model error convergence to zero, we define the following formulation,
\begin{alignat}{2}
\label{eq.errorconvergence}
&\frac{{de\left( t \right)}}{{dt}} =  - \mu \exp \left( t \right)\Psi \left( {e\left( t \right)} \right)\\
&\Psi \left( {e\left( t \right)} \right) = \frac{{\left( {1 + \exp ( - \delta )} \right)\left( {1 - \exp ( - \delta {e_i}(t))} \right)}}{{\left( {1 - \exp ( - \delta )} \right)\left( {1 + \exp ( - \delta {e_i}(t))} \right)}} \nonumber
\end{alignat}
where $\mu>0$ is the constant that can adjust the convergence rate; ${{\Psi }\left( {{e_i}(t)} \right)}$ denotes the activation function, and $\xi  \ge 2$, which makes $0 \le \left| {{e_i}(t)} \right| \le 1$1. Clearly, the error in (\ref{eq.errorconvergence}) converges  to zero with exponential convergence.

The function in (\ref{eq.errorconvergence})  is expanded as
\begin{alignat}{2}
&B\left( t \right)\dot l \left( t \right) =  - \dot B\left( t \right)l\left( t \right) + \dot P(t)\nonumber\\
&- \mu \exp \left( t \right) {\Psi }\left(l(t)-l^b(t)+{B(t)l(t) - P(t)} \right) 
\label{eq.novel VP-ZNN1}
\end{alignat}

We further modify the VP-ZNN in (\ref{eq.novel VP-ZNN1}) as
\begin{alignat}{2}
\dot l\left( t \right) &= \left( {I - B\left( t \right)} \right)\dot l\left( t \right)-\dot{B}(t)l(t) + \dot P\left( t \right)\nonumber \\
&- \mu \exp \left( t \right) {\Psi }\left( l(t)-l^b(t)+{B\left( t \right)l\left( t \right) - P\left( t \right)} \right) 
\end{alignat}

For comparison,  the traditional gradient descent-based recurrent neural network is denoted as
\begin{alignat}{2}
\dot l\left( t \right) = \mu \left( { - l(t) + {P_\Omega }\left[ {l(t) - \left( {B(t)l(t) - P(t)} \right)} \right]} \right)
\end{alignat}


\begin{figure}[htp!]
\centering{
\includegraphics[width=0.99\linewidth]{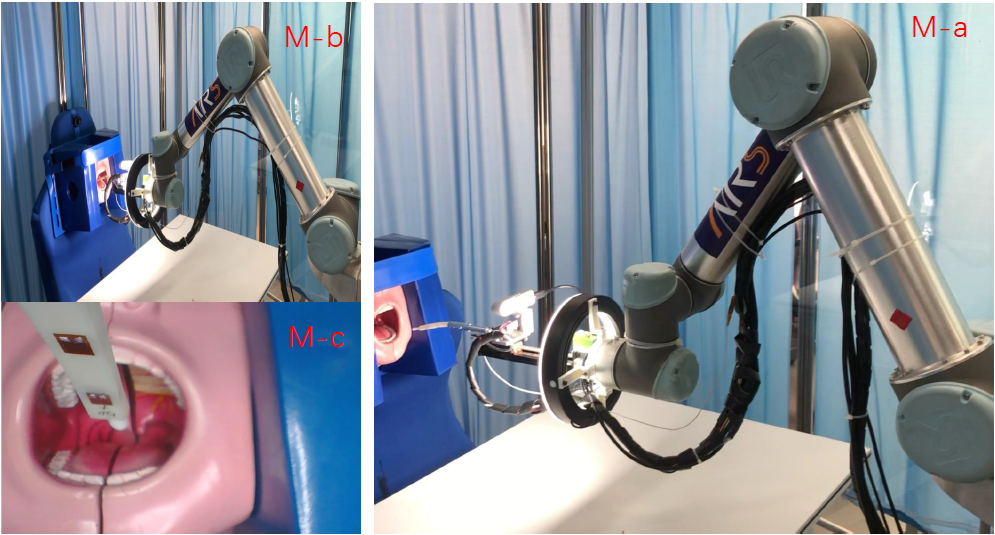}}
\caption{Demonstration with an oral cavity phantom. In our experiments, the 1:1 human  oral  cavity is tested.}
\label{fig.phantom}
\end{figure}

\begin{figure*}[htp!]
\centering{
\includegraphics[width=0.7\linewidth, height=0.4 \linewidth]{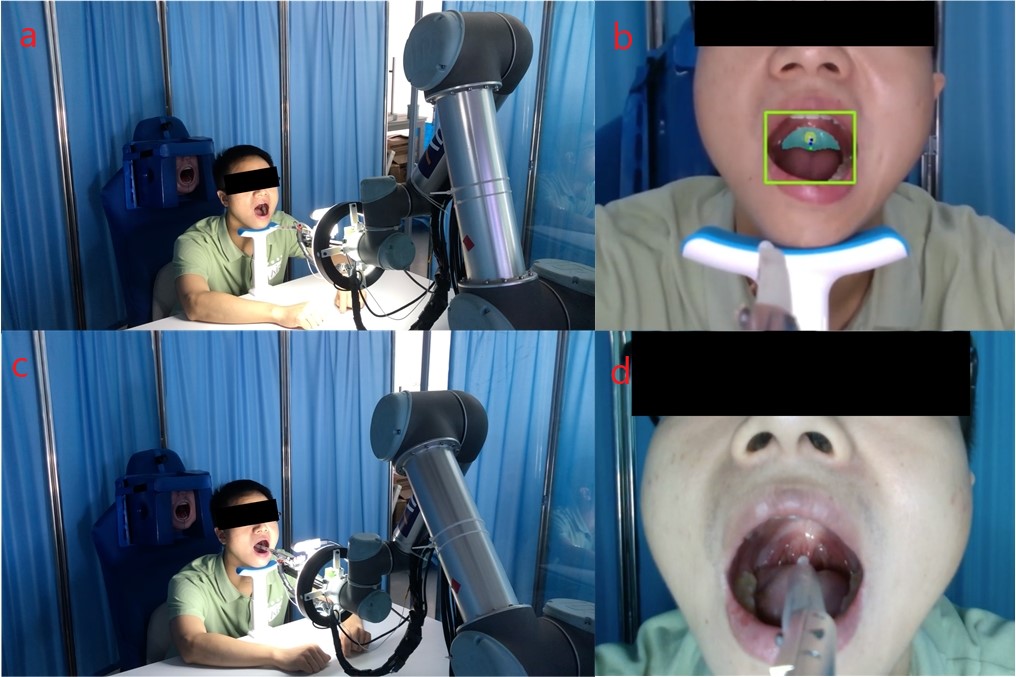}}
\caption{ 
Sampling process from vision detection to sampling tasks. In our experiments, many volunteers were tested, and the recognition rate,
 control precision, sampling time, and sampling contact force were recorded.}
\label{fig.experiment}
\end{figure*}

\begin{figure*}
\centering
\subfigure[Motion trajectories.]{\label{fig.motiontraj}
\includegraphics[width=0.3\linewidth]{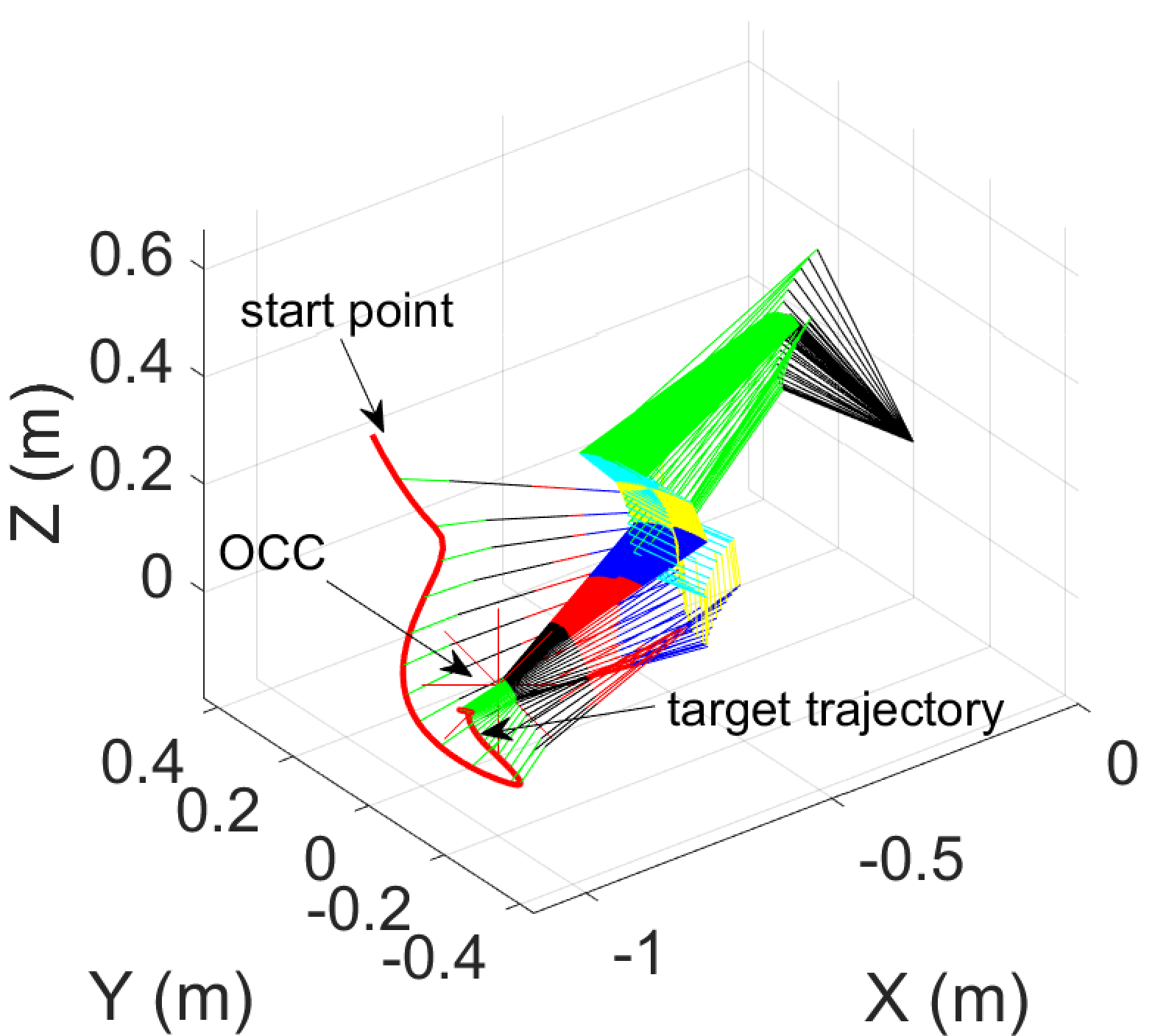}}
\subfigure[Joint trajectories.]{\label{fig.Joint}
\includegraphics[width=0.35 \linewidth]{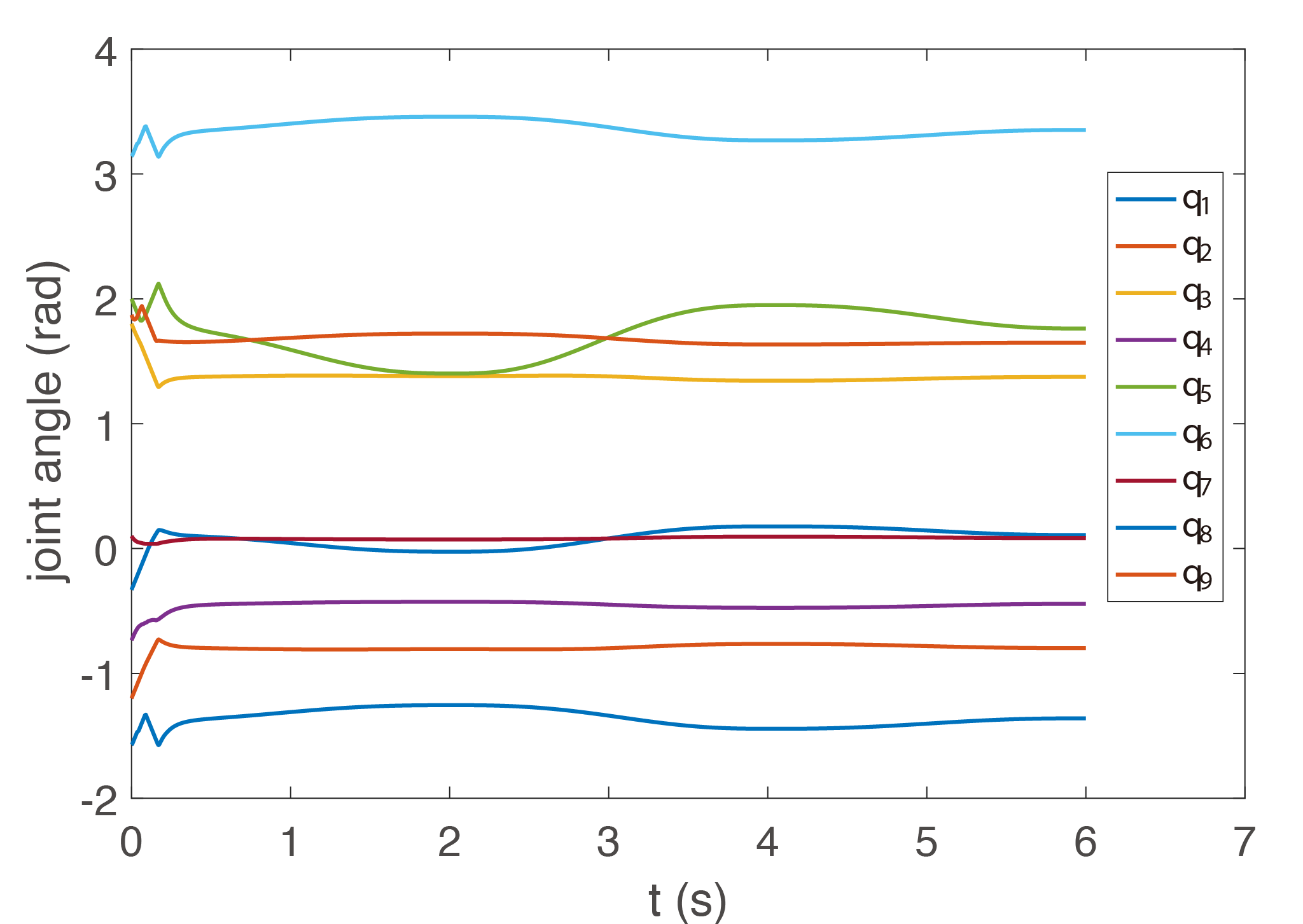}}

\subfigure[Joint angle velocity.]{\label{fig.Jointvel}
\includegraphics[width=0.35 \linewidth]{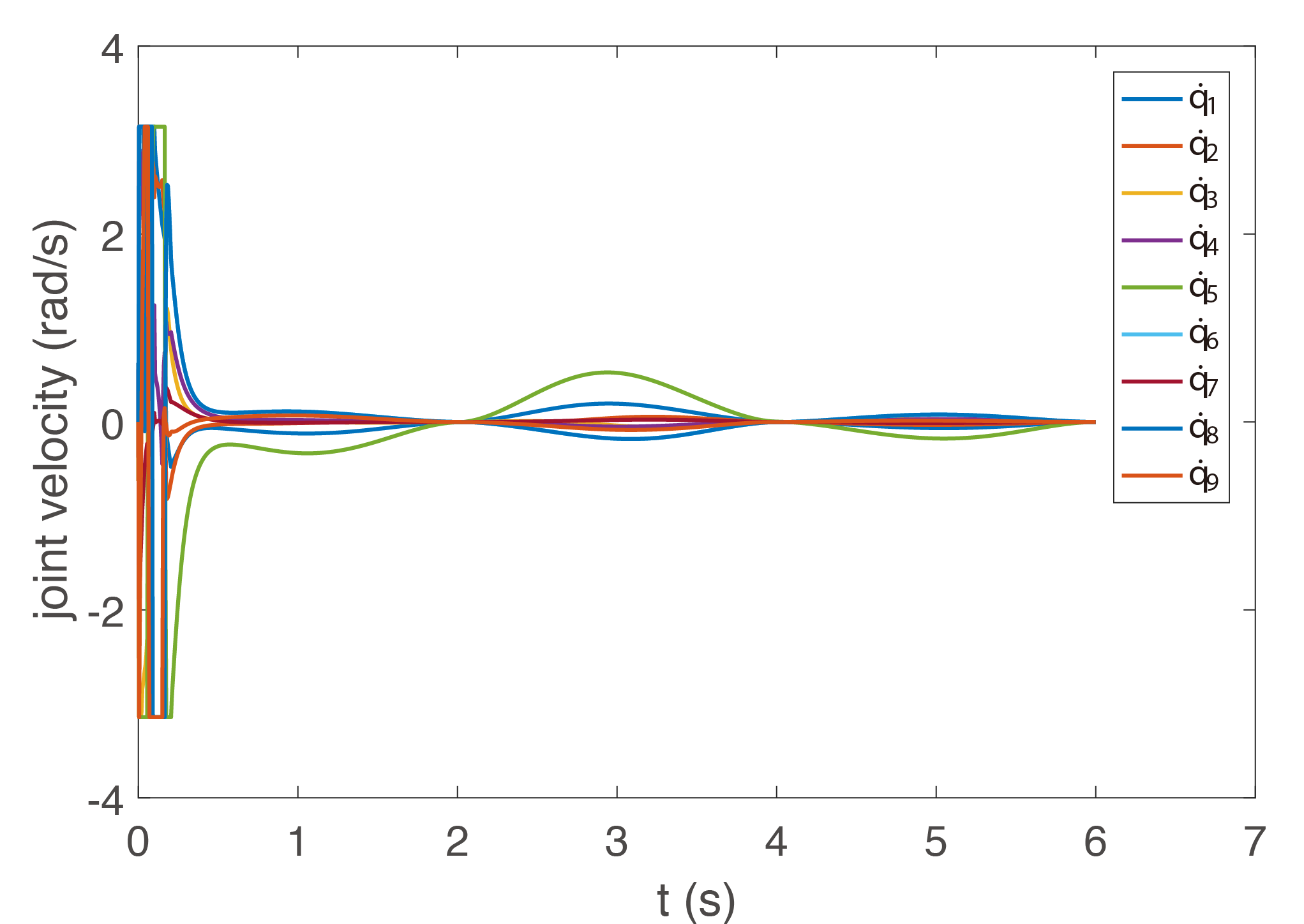}}
\subfigure[Tracking errors.]{\label{fig.trackingerror}
\includegraphics[width=0.35 \linewidth]{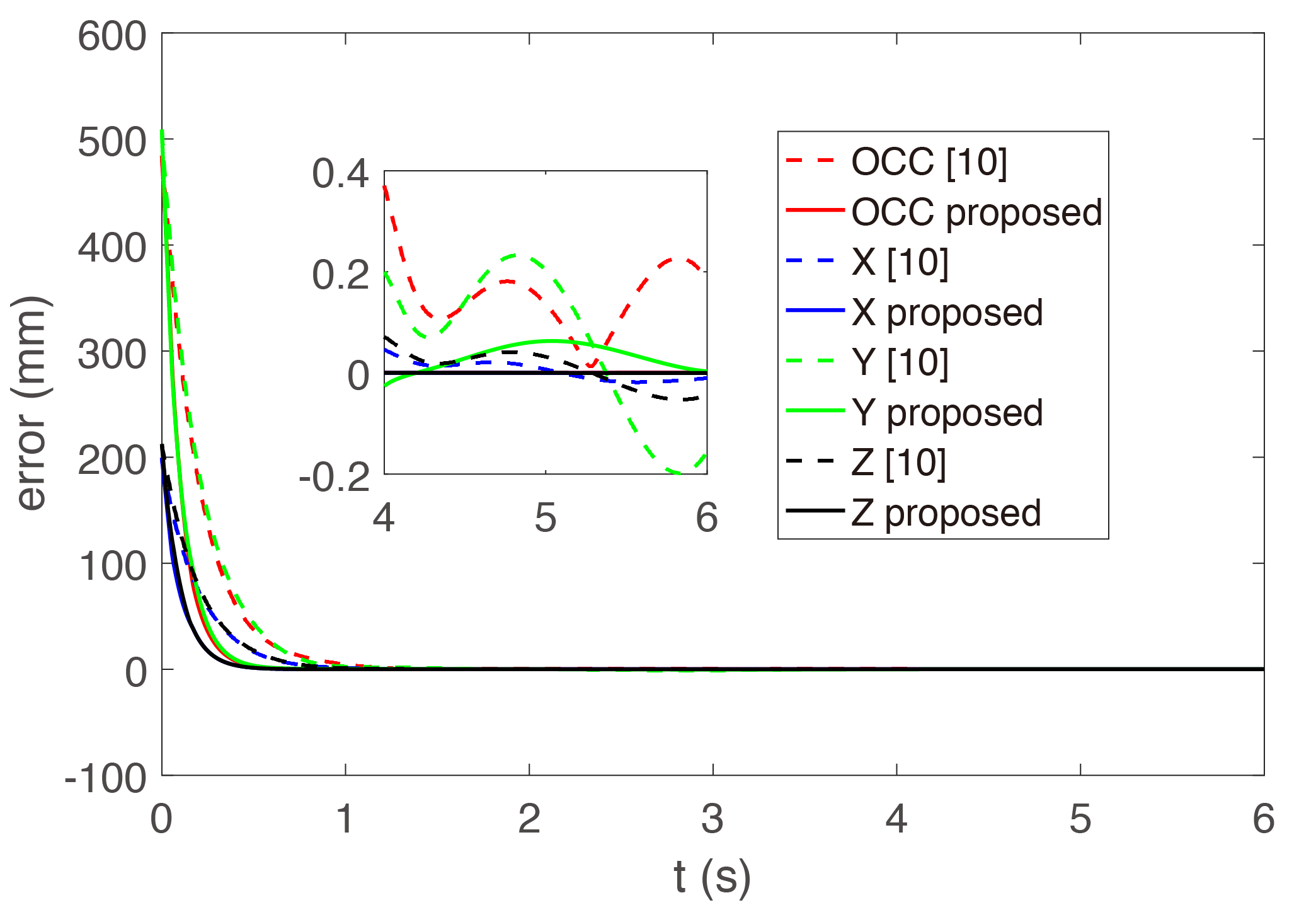}}
\caption{ 
Tracking results: the manipulator track of the desired trajectories (sampling tasks) in the Cartesian space. (a) The 3D trajectories of the manipulator; (b) the joint angle trajectories; (c) the joint velocity trajectories (located in bounds); and (d) comparison experiments with \cite{su2020improved}.}
\label{fig.trackingresults}
\end{figure*}




\section{Experiments} 
\label{sec.experiments}

In this section, we present the tests with an oral cavity phantom and volunteers using the OP-swab robot system, approval from the Institutional Review Board of The Chinese University of Hong Kong, Shenzhen was obtained(IRB number CUHKSZ-D-20210002). Then, the advantages and disadvantages are summarized and discussed. The parameters of the VP-ZNN are set as: $\varepsilon_0=0.1$, $\varepsilon_1=10$, $\varepsilon_2=10$, $\mu=0.01$, $k_1=10$, and $k_2=10$. 

\subsection{Demonstration with Phantom Experiments}
\label{subsec.phantom}

For safety reasons, first, we conduct the experiments with an oral cavity phantom, which will help to confirm the safety of the OP-swab robot system. The oral cavity phantom has a 1:1 size of human oral cavity, which enables natural simulation of workflow across OP-swab sampling and allows us to work extremely close to conditions in practice. The procedures of experiments are as follows:
\begin{itemize}

   \item  The phantom oral cavity is detected and segmented with the RealSense Camera, which is configured on the RFC manipulator. Because the sampling areas are the left and right tonsils and the palate area, we recognize and locate the target position by Mask R-CNN.

    \item The desired sampling trajectories are obtained by online motion planning with an oral cavity constraint, in which we locate the target position in the first step and then generate the trajectories. The desired trajectories are a piecewise straight line from the palate area to the left tonsils to the right tonsils in the Cartesian space, where the relationship between each axis and time is the minimum jerk curve.

    \item The robot is driven to perform the sampling tasks. In this phase, the desired joint trajectories are obtained from optimization control methods. Moreover, multiple protection mechanisms with force sensing are activated during robot operation.

\end{itemize}

The experimental scenarios of the oral cavity phantom are shown in Fig.~\ref{fig.phantom}.
We collect the dataset for the oral cavity phantom which is trained by Mask R-CNN \cite{he2017mask}. 
To obtain the category and location of the oral sampling areas, an oral visual detector is trained and obtained. First, an oral cavity image dataset is created by collecting from RealSense D435i. Second, the Mask R-CNN with a Inception v4 module as its backbone is trained on the oral cavity image dataset and an oral cavity detector is obtained. Finally, oral cavity coordinates are mapped to depth images to obtain the depth values of the oral cavity. We achieved a 95\% recognition success rate on the model, and the sampling time is less than 20 s.

\subsection{Demonstration with Volunteers Experiments}
On the basis of many trials of the OP-swab RFC robot system with the oral cavity phantom, the robustness and safety were significantly improved. Then, we conducted a number of volunteer experiments and achieved milestone significance for the future clinical application of OP-swab sampling. The procedures of experiments are the same as those described in Sec.~\ref{subsec.phantom}. The dataset of the human oral cavity is collected by 29 volunteers and trained using Mask R-CNN.

\begin{figure}[htp!]
\centering{
\includegraphics[width=0.85\linewidth]{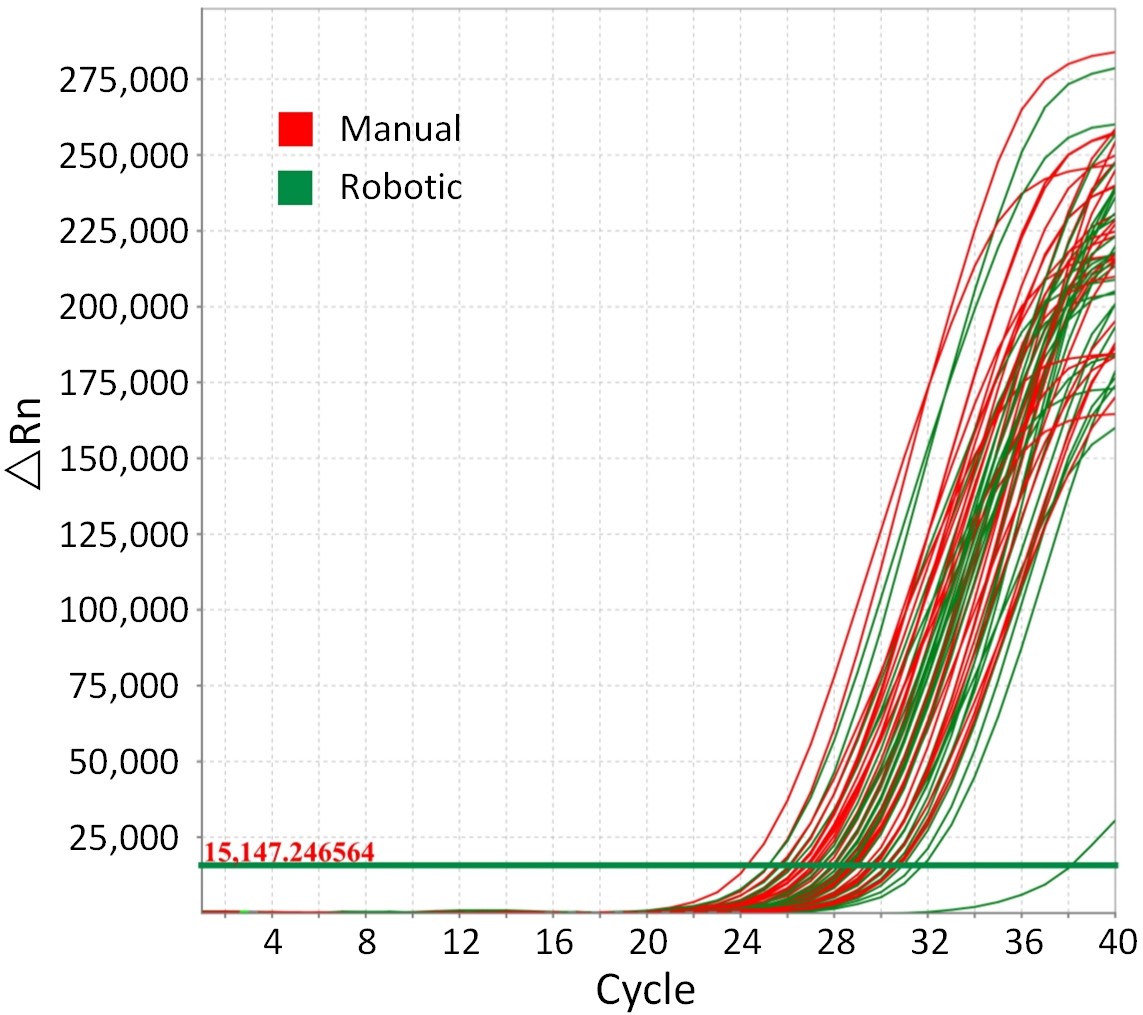}}
 \caption{ RT-PCR test results.}
\label{fig.PRC_comparison}
\end{figure}

The experimental scenarios of different subjects are shown in Fig.~\ref{fig.experiment}. The desired trajectories are the same as in Sec.~\ref{subsec.phantom}. We design the piecewise straight-line trajectories from the palate to the right tonsils to the left tonsils in the Cartesian space, and the relationship between each axis and time is the minimum jerk curve. Figure~\ref{fig.trackingresults} shows the tracking results associated with one of the trials of planning trajectories. The motion trajectories in the Cartesian space are shown in Fig.~\ref{fig.motiontraj}. Figures~\ref{fig.Joint} and \ref{fig.Jointvel} show that the joint trajectories located in the physical limits and joint velocity are smooth within the velocity limits, respectively. Figure~\ref{fig.trackingerror} shows the tracking error in the Cartesian space, where all errors rapidly converge to a small value (0.02 mm). In addition, we added a comparison experiment shown in Fig.~\ref{fig.trackingerror}, where the proposed method has a faster convergence rate and smaller errors than \cite{su2020improved}. The average recognition rate, control precision, sampling time, and sampling contact force are recorded and shown in Table.~\ref{tab.samplingparameters}. The comparative experiments were conducted to examine the effectiveness of robotic OP-swab sampling. The swab quality was verified according to the threshold cycle (Ct) value of the selected reference gene (RNase P) by the RT-PCR test ~\cite{radonic2004guideline}. Figure ~\ref{fig.PRC_comparison} shows the RT-PCR test results compared with those for the manual scheme. OP-swab with Ct values of $\leq 37$ and $> 37$ are considered as qualified and unqualified samples, respectively, and the details are shown in Table ~\ref{tab.PCRparameters}. The test results show that the samples are qualified.

 \begin{table}[ht]
\centering
\setlength{\tabcolsep}{3.5mm}
\caption{Average sampling parameters.}
\begin{tabular}{|l|c|c|}
\hline
\diagbox{Metric }{Types} & Phantom & Volunteers\\  
\hline
Recognition rate & 95\% & 93\%  \\ 
\hline
Control precision & $\le $0.2 mm & $\le$0.2 mm\\ 
\hline
Sampling time & 18 $\pm$ 2 s  & 20 $\pm$ 4 s\\
\hline
Sampling force & $\approx$ 150 mN & $\approx$ 150 mN\\ 
\hline
\end{tabular}
\label{tab.samplingparameters}
\end{table}
 



\begin{table}[ht]
\centering
\setlength{\tabcolsep}{1.5mm}
\caption{Ct distribution.}
\begin{tabular}{|l|c|c|c|c|c|c|}
\hline
Ct &  (24–27)& (27–30)& (30–33)& (33–37)&\textgreater 37 & Qualified rate\\
\hline
Manual  & 11 & 16& 3 & 0 &0 & 100\%  \\ 
\hline
Robotic  & 5 & 21& 3 &0 & 1 & 96.67\%\\ 
\hline

\end{tabular}
\label{tab.PCRparameters}
\end{table}

\section{Conclusion and  Future Work} \label{sec.conclusion}
In this study, we designed a 9-DOF redundant RFC robot to assist the COVID-19 OP-swab sampling. Moreover, we formulate the sampling tasks, physical limits, and OCC constraints as a novel optimization problem. Then, a VP-ZNN is proposed to solve the multi-constraint optimization problem online. The experimental results of the oral cavity phantom and volunteer experiments demonstrate the effectiveness of the designed robot system and proposed control methods. The average sampling time on phantoms and volunteers are 18 s and 20 s respectively. 
To maintain the sterility of the arm and effector between patients, the MPA is designed as a disposable device with a quick connector that is easily to be replaced, where the disposable protective film is attached to the MPA.
In the future, we will focus on improving the robustness of the system and move forward to clinical testing. 

\appendices
\section*{Acknowledgment}
The authors would like to thank the Prof. Xi Zhu, Mr. Yi Liang, Ms. Kerui Yi and Mr. Xiaochuan Lin for their suggestions and engineering contributions to this work. This work was supported by the Shenzhen Fundamental Research grant (JCYJ20180508162406177, JCYJ20190806142818365) and the National Natural Science Foundation of China (U1613216, 62006204, 61903100) from The Chinese University of Hong Kong, Shenzhen. This work was also partially supported by the Shenzhen Institute of Artificial Intelligence and Robotics for Society.  


\bibliographystyle{IEEEtran}
\bibliography{references}

\begin{thebibliography}{10}
\providecommand{\url}[1]{#1}
\csname url@samestyle\endcsname
\providecommand{\newblock}{\relax}
\providecommand{\bibinfo}[2]{#2}
\providecommand{\BIBentrySTDinterwordspacing}{\spaceskip=0pt\relax}
\providecommand{\BIBentryALTinterwordstretchfactor}{4}
\providecommand{\BIBentryALTinterwordspacing}{\spaceskip=\fontdimen2\font plus
\BIBentryALTinterwordstretchfactor\fontdimen3\font minus
  \fontdimen4\font\relax}
\providecommand{\BIBforeignlanguage}[2]{{%
\expandafter\ifx\csname l@#1\endcsname\relax
\typeout{** WARNING: IEEEtran.bst: No hyphenation pattern has been}%
\typeout{** loaded for the language `#1'. Using the pattern for}%
\typeout{** the default language instead.}%
\else
\language=\csname l@#1\endcsname
\fi
#2}}
\providecommand{\BIBdecl}{\relax}
\BIBdecl

\bibitem{wang2020comparison}
X.~Wang, L.~Tan, X.~Wang, W.~Liu, Y.~Lu, L.~Cheng, and Z.~Sun, ``Comparison of
  nasopharyngeal and oropharyngeal swabs for sars-cov-2 detection in 353
  patients received tests with both specimens simultaneously,''
  \emph{International Journal of Infectious Diseases}, 2020.

\bibitem{xu2020open}
B.~Xu, M.~U. Kraemer, B.~Gutierrez, S.~Mekaru, K.~Sewalk, A.~Loskill, L.~Wang,
  E.~Cohn, S.~Hill, A.~Zarebski \emph{et~al.}, ``Open access epidemiological
  data from the covid-19 outbreak,'' \emph{The Lancet Infectious Diseases},
  vol.~20, no.~5, p. 534, 2020.

\bibitem{hindson2020covid}
J.~Hindson, ``Covid-19: faecal--oral transmission?'' \emph{Nature Reviews
  Gastroenterology \& Hepatology}, vol.~17, no.~5, pp. 259--259, 2020.

\bibitem{yang2020combating}
G.-Z. Yang, B.~J. Nelson, R.~R. Murphy, H.~Choset, H.~Christensen, S.~H.
  Collins, P.~Dario, K.~Goldberg, K.~Ikuta, N.~Jacobstein \emph{et~al.},
  ``Combating covid-19—the role of robotics in managing public health and
  infectious diseases,'' 2020.

\bibitem{li2020clinical}
S.-Q. Li, W.-L. Guo, H.~Liu, T.~Wang, Y.-Y. Zhou, T.~Yu, C.-Y. Wang, Y.-M.
  Yang, N.-S. Zhong, N.-F. Zhang \emph{et~al.}, ``Clinical application of
  intelligent oropharyngeal-swab robot: Implication for covid-19 pandemic,''
  \emph{European Respiratory Journal}, 2020.

\bibitem{liu2020}
\BIBentryALTinterwordspacing
H.~Liu, ``Application of artificial intelligence robots in respiratory
  diseases,'' 2000. [Online]. Available:
  \url{https://wx.vzan.com/live/tvchat-76585257?ver=637228232672628287#/}
\BIBentrySTDinterwordspacing

\bibitem{wang2020design}
S.~Wang, K.~Wang, H.~Liu, and Z.~Hou, ``Design of a low-cost miniature robot to
  assist the covid-19 nasopharyngeal swab sampling,'' \emph{arXiv preprint
  arXiv:2005.12679}, 2020.

\bibitem{Danmark2020}
``Danish startup develops throat swabbing robot for covid-19 testing,''
  \url{https://www.therobotreport.com/danish-startup-develops-throat-swabbing-robot-for-covid-19-testing/}.

\bibitem{li2015novel}
Z.~Li, J.~Feiling, H.~Ren, and H.~Yu, ``A novel tele-operated flexible robot
  targeted for minimally invasive robotic surgery,'' \emph{Engineering},
  vol.~1, no.~1, pp. 073--078, 2015.

\bibitem{su2020improved}
H.~Su, Y.~Hu, H.~R. Karimi, A.~Knoll, G.~Ferrigno, and E.~De~Momi, ``Improved
  recurrent neural network-based manipulator control with remote center of
  motion constraints: Experimental results,'' \emph{Neural Networks}, vol. 131,
  pp. 291--299, 2020.

\bibitem{su2019improved}
H.~Su, C.~Yang, G.~Ferrigno, and E.~De~Momi, ``Improved human--robot
  collaborative control of redundant robot for teleoperated minimally invasive
  surgery,'' \emph{IEEE Robotics and Automation Letters}, vol.~4, no.~2, pp.
  1447--1453, 2019.

\bibitem{mirrazavi2018unified}
S.~S. Mirrazavi~Salehian, N.~Figueroa, and A.~Billard, ``A unified framework
  for coordinated multi-arm motion planning,'' \emph{The International Journal
  of Robotics Research}, vol.~37, no.~10, pp. 1205--1232, 2018.

\bibitem{wang2017prestressed}
Z.~Wang, Y.~Torigoe, and S.~Hirai, ``A prestressed soft gripper: design,
  modeling, fabrication, and tests for food handling,'' \emph{IEEE Robotics and
  Automation Letters}, vol.~2, no.~4, pp. 1909--1916, 2017.

\bibitem{li2017asymmetric}
Z.~Li, B.~Huang, A.~Ajoudani, C.~Yang, C.-Y. Su, and A.~Bicchi, ``Asymmetric
  bimanual control of dual-arm exoskeletons for human-cooperative
  manipulations,'' \emph{IEEE Transactions on Robotics}, vol.~34, no.~1, pp.
  264--271, 2017.

\bibitem{li2018adaptive}
Z.~Li, J.~Li, S.~Zhao, Y.~Yuan, Y.~Kang, and C.~P. Chen, ``Adaptive neural
  control of a kinematically redundant exoskeleton robot using brain--machine
  interfaces,'' \emph{IEEE transactions on neural networks and learning
  systems}, vol.~30, no.~12, pp. 3558--3571, 2018.

\bibitem{facchinei2014solving}
F.~Facchinei, C.~Kanzow, and S.~Sagratella, ``Solving quasi-variational
  inequalities via their kkt conditions,'' \emph{Mathematical Programming},
  vol. 144, no. 1-2, pp. 369--412, 2014.

\bibitem{he2017mask}
K.~He, G.~Gkioxari, P.~Doll{\'a}r, and R.~Girshick, ``Mask r-cnn,'' in
  \emph{Proceedings of the IEEE international conference on computer vision},
  2017, pp. 2961--2969.

\bibitem{radonic2004guideline}
A.~Radoni{\'c}, S.~Thulke, I.~M. Mackay, O.~Landt, W.~Siegert, and A.~Nitsche,
  ``Guideline to reference gene selection for quantitative real-time pcr,''
  \emph{Biochemical and biophysical research communications}, vol. 313, no.~4,
  pp. 856--862, 2004.

\end{thebibliography}



\end{document}